\documentclass[sigconf]{acmart}
\copyrightyear{2025}
\acmYear{2025}
\setcopyright{acmlicensed}
\acmConference[ICAIF '25]{6th ACM International Conference on AI in Finance}{November 15--18, 2025}{Singapore, Singapore}
\acmBooktitle{6th ACM International Conference on AI in Finance (ICAIF '25), November 15--18, 2025, Singapore, Singapore}
\acmDOI{10.1145/3768292.3770398}
\acmISBN{979-8-4007-2220-2/2025/11}

\usepackage{subcaption}
\usepackage{multirow}
\usepackage{makecell}

\newcommand\tcapfig[1]{\captionsetup{position=top, font=normalsize, labelfont=bf, textfont=normalfont, justification=centering, margin=0mm, aboveskip=2mm, belowskip=0mm, labelsep=colon, singlelinecheck=false}\caption{#1}}
\newcommand\bnotefig[2][\linewidth]{%
  \captionsetup{position=bottom, font=footnotesize, textfont=normalfont, margin=1mm, skip=2mm, justification=justified, singlelinecheck=false}%
  \begin{minipage}{#1}
    \caption*{#2}
  \end{minipage}
}

\begin{document}

\title{Attention Factors for Statistical Arbitrage}

\author{Elliot L. Epstein}
\affiliation{%
  \institution{Stanford University}
  \city{Stanford}
  \country{United States}}
\email{epsteine@stanford.edu}

\author{Rose Wang}
\affiliation{%
  \institution{Stanford University}
  \city{Stanford}
  \country{United States}}
\email{rosew47@stanford.edu}

\author{Jaewon Choi}
\affiliation{%
  \institution{Hanwha Life}
  \city{Seoul}
  \country{South Korea}}
\email{jaewonch@hanwha.com}

\author{Markus Pelger}
\affiliation{%
  \institution{Stanford University}
  \city{Stanford}
  \country{United States}}
\email{mpelger@stanford.edu}

\begin{abstract}

Statistical arbitrage exploits temporal price differences between similar assets. We develop a framework to \emph{jointly} identify similar assets through factors, identify mispricing and form a trading policy that maximizes risk-adjusted performance after trading costs. Our \emph{Attention Factors} are conditional latent factors that are the most useful for arbitrage trading. They are learned from firm characteristic embeddings that allow for complex interactions. We identify time-series signals from the residual portfolios of our factors with a general sequence model. 
Estimating factors and the arbitrage trading strategy jointly is crucial to maximize profitability after trading costs. 
In a comprehensive empirical study we show that our Attention Factor model achieves an out-of-sample Sharpe ratio above 4 on the largest U.S. equities over a 24-year period. Our one-step solution yields an unprecedented Sharpe ratio of 2.3 net of transaction costs. We show that weak factors are important for arbitrage trading.

\end{abstract}


\maketitle
\section{Introduction}

Statistical arbitrage exploits temporal price differences between similar assets using statistical methods. Conceptually, these methods are based on relative trades between a stock and a mimicking portfolio. The mimicking portfolio is constructed to be ``similar'' to the target stock, usually based on historical co-movements in the price time-series. When the spread between the prices of the two comparison assets widens, the arbitrageur sells the winner and buys the loser. If their prices move back together, the arbitrageur will profit. Statistical arbitrage trading has to solve the following three key problems: Given a large universe of assets, what are long-short portfolios of similar assets? Given these portfolios, what are time-series signals that indicate the presence of temporary price deviations? Lastly, given these signals, how should an arbitrageur trade them to maximize risk-adjusted performance after trading costs? Each of these three questions poses substantial challenges, that prior work has only partly addressed.

Previous approaches have studied this problem as a two-step approach, where the first step identifies similar assets separately from the trading objective. Similarity between assets can be captured by similar exposure to risk factors. Arbitrage portfolios are trades relative to mimicking stock portfolios with the same exposure to risk factors. A common approach is to use principal component analysis (PCA) factors, which maximize the explained correlation in a panel and where mimicking portfolios are assets with high correlation with the target stocks. The second step in arbitrage trading is to identify time-series signals from the residuals of a candidate factor model and form a trading policy. The leading approach is \citep{dlsa}, which outperforms the benchmarks in this literature. It uses PCA-type factors in the first step and a general sequence model in the second step. It achieves high Sharpe ratios before trading costs, but degrading Sharpe ratios after trading costs. The key issue in a two-step approach is that the factors cannot adjust to reduce trading costs for arbitrage strategies. For example, PCA factors have high turnover and large short positions, which diminish net performance. We provide a solution with our one-step approach.

In this paper, we propose the \emph{Attention Factor Model}, a framework that \emph{jointly} learns tradable arbitrage factors and arbitrage portfolio allocations in a computationally efficient manner. Our Attention Factors are conditional latent factors. The estimation objective is not to explain variation, but to construct profitable arbitrage strategies after trading costs. The attention mechanism learns embeddings of firm characteristics and allows to capture general dependencies of the factors on firm characteristics with complex interactions. A general sequence model learns time-series signals from the residuals of our attention factors with the joint objective of maximizing the net Sharpe ratio and explained variance. Figure \ref{fig:banner} illustrates the conceptual structure. 

We perform a comprehensive empirical out-of-sample analysis on 24 years of daily returns of the 500 largest and most liquid U.S. equities using an extensive set of firm characteristics. The Attention Factor model achieves an annualized Sharpe ratio above 4 without trading frictions and 2.3 with trading frictions, significantly outperforming prior work with an 84\% increase in net Sharpe ratio over the current state-of-the-art model in ~\citep{dlsa}. Our arbitrage strategy yields an annual return of 16\% while being uncorrelated to market risk. The Attention Factors have an interpretable structure, where the loadings are closely related to industry sectors. Our study provides evidence for weak factors that explain less variation but are important for identifying temporal mispricing. 

 \begin{figure*}
    \centering
    \tcapfig{Conceptual Attention Factor Model}
    \includegraphics[width=1\linewidth]{figures/attn_factors.pdf}
    \bnotefig{The figure illustrates the conceptual structure of the Attention Factor model. Left: Attention factors are constructed by computing scaled inner products between embedded characteristics for each asset and the $K$ query vectors $Q_k$. Right: The statistical arbitrage methodology. First, for each asset, a replicating portfolio based on the attention factors is created, giving a residual mispricing. Second, a series of lagged residuals are used to construct the portfolio weights in the residual space, using a Long Convolution model for sequence modeling. Finally, the portfolio weights are mapped back to the asset space via a composition matrix, giving the next-period portfolio return. }
    \label{fig:banner}
\end{figure*}

\section{Related Work}
\label{sec:related}

\paragraph{Statistical Arbitrage}

Our paper builds on the classical statistical arbitrage literature, in which the three main problems of residual portfolio construction, time-series pattern extraction, and allocation decision have traditionally been considered independently. Classical statistical methods of generating arbitrage portfolios use parametric methods and have mostly focused on obtaining multiple pairs or small portfolios of assets, using techniques like the distance method of \citep{10.1093/rfs/hhj020}, the cointegration approach of \citep{Vidyamurthy}, or copulas as in \citep{Copula}. In contrast, more general methods that exploit large panels of stock returns include the use of PCA factor models, as in \citep{avelee} and its extension in \citep{papayeo}, and the maximization of mean-reversion and sparsity statistics as in \citep{dAspremont}. Alternative parametric models include \citep{,orig-stoch-opt,Cartea2016,Tourin2016}. The most closely related paper to our work is \citep{dlsa}, which uses transformer models to extract general time-series patterns from residuals. The residuals are obtained from PCA and IPCA factors to optimize explained variation. Our approach unifies factor extraction and residual trading within a single learning objective.

\paragraph{Machine Learning in Asset Pricing}
Our paper is complementary to the fast growing literature that uses machine learning methods for asset pricing. While the asset pricing literature aims to explain the risk premia of assets, our focus is on the residual component which is not explained by the asset pricing models. \citep{dlap,bryzgalova2019,kozak_shrinking_2020,NBERw33351} estimate the stochastic discount factor (SDF), which explains the risk premia of assets, with deep neural networks, decision trees, elastic net regularization and attention methods. \citep{gu2018,bianchi2019,freyberger2017dissecting,Turan_et_al2022,KANIEL202394,DeMiguel_et_al2022} predict asset returns with machine learning methods.

\paragraph{Statistical Factor Modeling}
The workhorse models in equity asset pricing are based on linear factor models exemplified by \citep{FAMA19933,FAMA20151}. Recently, new methods have been developed to extract statistical factors from large panels with various versions of PCA that explain the systematic comovement between assets \citep{BaiNg2002,Fan2013}. Motivated by Arbitrage Pricing Theory (APT), systematic risk factors are expected to explain the cross-section of expected returns. Extensions of PCA include RP-PCA \citep{Lettau_Pelger_RFS} to account for pricing errors, state-dependent factors in \citep{Pelger03072022}, interpretable PCA \citep{Pelger02102022}, high-frequency PCA \citep{Pelger2020}, and conditional factor models in Instrumented PCA (IPCA) \citep{KELLY2019501} linking latent loadings to observable characteristics. Statistical factors that explain the variation in panels are complementary to our work as they have a different objective. Our method estimates factors that are the most useful for arbitrage trading.

\paragraph{Machine learning for Time-Series}
Our paper builds on the literature for time-series modeling with sequence models, which typically solve a time-series prediction problem. We estimate a LongConv~\citep{fu2023simplehardwareefficientlongconvolutions} model jointly with our Attention Factor model ~\citep{vaswani_attention_2023} with a trading objective. Popular sequence models to learn general time-series patterns are Transformer~\citep{vaswani_attention_2023} models and S4~\citep{gu2022efficiently} models. The Set-Sequence Model~\citep{epstein2025set} captures joint dependencies for arbitrary sequence models. Low-rank Gaussian copula processes model joint distributions with tractable structure \citep{salinas2019highdimensionalmultivariateforecastinglowrank}, while global–local networks exploit parameter sharing with series-specific conditioning \citep{NEURIPS2019_3a0844ce}. Transformer-based models dominate recent benchmarks: Crossformer introduces cross-dimension attention for multivariate dependencies \citep{zhang2023crossformer}, iTransformer inverts tokenization to attend over variates and scales to long horizons \citep{liu_itransformer_2024}, and S4-based models use structured state spaces to efficiently handle long sequences \citep{gu2022efficiently}.

 \section{Method}

 \paragraph{Notation}
We consider $N$ assets with returns $R_t \in \mathbb{R}^N$ with $M$ time-varying characteristics $X_t \in \mathbb{R}^{N \times M}$ for the times $t=1,\dots,T$.

 \paragraph{Problem} The fundamental problem of statistical arbitrage consists of three elements: (1) Identification of similar assets to generate arbitrage portfolios, (2) extraction of time-series signals for temporary deviations of similarity between assets and (3) a trading policy in the arbitrage portfolios based on the time-series signals. We provide a general end-to-end solution for each element.

 \subsection{Factor Model}
 
\paragraph{Conditional Factor Model} 
Factor models explain the returns of a cross-section of assets in terms of their exposure to factors $F_t = (F_{1,t}, \dots, F_{K,t})$. We use factors to identify similar assets, where similarity is defined as the same exposure to factors. We assume that asset returns can be modeled by a conditional factor model:
$$R_{i,t} = \beta_{i,t-1}^T F_t + \epsilon_{i,t}, \qquad \text{$t=1,...,T$ and $i=1,...N$.}$$
The $K$ tradable factors $F_t \in \mathbb{R}^K$ capture systematic risk, while the loadings $\beta_{i,t-1}$ are time-varying and based on information up to time $t-1$. This general formulation includes the empirically most successful factor models. 

Through the lens of Arbitrage Pricing Theory (APT) \citep{Lettau_Pelger_RFS}, if the factors capture all relevant sources of systematic risk, then the factor portfolio $\beta_{i,t-1}^T F_t$ is the ``fair price'', and the residual portfolio, $\epsilon_{i,t}$, given by
$$\epsilon_{i,t} = R_{i,t} -\beta_{i,t-1}^T F_t,$$
identifies mispricing. Arbitrage trading aims to exploit temporal patterns in the residuals. 

\paragraph{Candidate Factors}
Empirically successful factor models include observed fundamental factors and statistical factors. Examples of fundamental factors are the market factor in the CAPM model, and the Fama-French 3- and 5-factor models ~\citep{FAMA19933} that include a market, size, value, respectively, investment and profitability factor. The Fama-French factors are tradeable portfolios, $F_t=\omega^{\text{FF}}_{t-1} R_t$, where portfolio weights $\omega^{\text{FF}}_{t-1}$ depend on past firm characteristics like size or book-to-market ratios. Statistical factor models encompass unconditional and conditional factor models. The most widely used unconditional factor models are based on versions of PCA; see \cite{Lettau_Pelger_RFS} for an overview. PCA estimates latent factors that explain the most cross-sectional variation. These latent factors are typically extracted as \( F_t = \omega^{\text{PCA}} R_t \), where \( \omega^{\text{PCA}}  \) are the eigenvectors of the top \( K \) eigenvalues of the return covariance matrix. Conditional statistical factors model the factor portfolio weights and loadings as functions of characteristics. A prominent example is IPCA in \citep{KELLY2019501} with $F_t=\omega^{\text{IPCA}}_{t-1} R_t$, where the weights $\omega^{\text{IPCA}}_{t-1}=X_{i,t-1}^{\top} B$ are a linear function of firm characteristics. This conditional structure allows exposures to evolve with firm attributes, linking the latent factors to observable fundamental information. However, the factors are still estimated to maximize explained variation in the cross-section.

\paragraph{Challenges for Arbitrage Trading}
The above factor models are not constructed with the objective to create profitable arbitrage portfolios. These models impose restrictive ad-hoc assumptions on the functional form of loadings and portfolio weights, and do not target factors based on a trading objective. Importantly, the resulting factor portfolios might have high trading costs in terms of turnover and shorting positions. We provide a solution that addresses all these challenges.

\subsection{Attention Arbitrage Factors}

 \paragraph{Residual Portfolios}
We estimate a conditional factor model that is optimal for arbitrage trading. As in factor models from the previous section, our factors are tradable portfolios
$$F_t = {\omega_{t-1}^{F}} R_t,$$
for a factor portfolio weight matrix $\omega_{t-1}^{\text{F}} \in \mathbb{R}^{K \times N}$. This implies that the 
residuals are traded portfolios as well
$$\epsilon_t = R_{t} -\beta_{t-1}^T F_t = R_t - \beta_{t-1}^T \omega_{t-1}^{\text{F}} R_t = \omega^{\epsilon}_{t-1} R_t,$$
for the implied projection matrix $\omega^{\epsilon}_{t-1} = I_{N} - \beta_{t-1}^T {\omega_{t-1}^{\text{F}}}.$

\paragraph{Attention Factors}
Our attention factors allow for a general functional form for the weights and loadings, that captures complex dependencies between characteristics. In our approach, for each point in time, the firm characteristics $X_t$ are first embedded as
\begin{equation*}
    \tilde{X}_{t} = X_t W^K, \quad W^K \in \mathbb{R}^{M \times d}.
\end{equation*}
Each asset is attended with a dot product to the query vector $Q_k \in \mathbb{R}^{d}$ for each factor, with query matrix $Q = (Q_1,\dots,Q_K)^T$ giving a factor weight matrix $\omega^{\text{F}}_{t-1}$ as
\begin{equation}
\label{attn_factors}
\omega^{\text{F}}_{t-1} = \operatorname{Softmax}( Q\tilde{X}^T_{t-1}  / \sqrt{d}),
\end{equation}
where Softmax is applied along each row to ensure that the resulting factors are normalized. The name attention factor is due to the similarity of Equation~\ref{attn_factors} with the multi-head attention mechanism used in Transformer~\citep{vaswani_attention_2023} models, but instead of standard attention that compares each token in a sequence (across time) to each other token, we compare each asset (in the cross section) with each factor. Simpler models, for example IPCA, are a special case of this formulation. 
Factor loadings $\beta_{t-1}$ and weights $\omega^{\text{F}}_{t-1}$ are mechanically related, as up to a rotation the loadings represent factor portfolio weights (see, for example, \citep{Ma2021}). We obtain the factor loadings as
$$\beta_{t-1}^T = {\omega_{t-1}^F}^T \left({\omega_{t-1}^F} ({\omega_{t-1}^F})^T + \lambda_{\text{ridge}} I_K \right)^{-1},$$ 
where we add a ridge penalty $\lambda_{\text{ridge}}$ for stability. Hence, the estimation of $\omega^{\text{F}}_{t-1}$ directly implies $\beta_{t-1}$ and $\omega^{\epsilon}_{t-1}$.

\begin{table*}[th]
\tcapfig{Firm Characteristics by Category}
\label{tab:firm_chars}
\small
\setlength{\tabcolsep}{4pt}  
\begin{tabular}{lllllll}
\toprule
      & \multicolumn{2}{l}{\textbf{Past Returns}}
      & \multicolumn{2}{l}{\textbf{Value}}
      & \multicolumn{2}{l}{\textbf{Investment}} \\
      & r2\_1        & Short-term momentum
      & A2ME         & Assets / market cap
      & Investment   & Investment \\
      & r12\_2       & Momentum
      & BEME         & Book-to-market ratio
      & NOA          & Net operating assets \\
      & r12\_7       & Intermediate momentum
      & C            & Cash + ST inv. / assets
      & DPI2A        & Change in PP\&E \\
      & r36\_13      & Long-term momentum
      & CF           & Free cash-flow / book value
      &              &  \\           
      & ST\_Rev      & Short-term reversal
      & CF2P         & Cash-flow / price
      &              &  \\
      & Ret\_D1      & Daily return
      & Q            & Tobin’s Q
      &              &  \\
      & Ret\_W1      & Weekly return
      & Lev          & Leverage
      &              &  \\
      & STD\_W1      & Weekly volatility
      & E2P          & Earnings / Price
      &              &  \\
\cmidrule(lr){2-3}\cmidrule(lr){4-5}\cmidrule(lr){6-7}
      & \multicolumn{2}{l}{\textbf{Trading Frictions}}
      & \multicolumn{2}{l}{\textbf{Profitability}}
      & \multicolumn{2}{l}{\textbf{Intangibles}} \\
      & AT           & Total assets
      & PROF         & Profitability
      & OA           & Operating accruals \\
      & LME          & Size
      & CTO          & Capital turnover
      & OL           & Operating leverage \\
      & LTurnover    & Turnover
      & FC2Y         & Fixed costs / sales
      & PCM          & Price-to-cost margin \\
      & Rel2High     & 52-week-high closeness
      & OP           & Operating profitability
      &              &  \\
      & Resid\_Var   & Residual variance
      & PM           & Profit margin
      &              &  \\
      & Spread       & Bid–ask spread
      & RNA          & Return on NOA
      &              &  \\
      & SUV          & Standard unexplained volume
      & D2A          & Capital intensity
      &              &  \\
      & Variance     & Variance
      &              &
      &              &  \\
      & Vol          & Weekly trading volume
      &              &
      &              &  \\
      & Beta         & Beta with market
      &              &
      &              &  \\
\bottomrule
\end{tabular}
\bnotefig[0.8\linewidth]{The table shows the 39 firm-specific characteristics (six categories) used as features to construct the attention factors. Construction details are in the Internet Appendix of \cite{dlap}.}
\end{table*}

\subsection{Arbitrage Trading}
 
 \paragraph{Arbitrage Portfolio}
The key idea of statistical arbitrage is to exploit predictable patterns in the time-series of residual portfolios. Traditionally, statistical arbitrage focuses on parametric mean-reversion patterns. We detect time-series patterns in the residual portfolios with a flexible data-driven filter based on convolutional networks using a trading objective. The arbitrage portfolio weight function depends on the time-series signals that we extract with LongConv~\citep{fu2023simplehardwareefficientlongconvolutions} from the past $s$ residuals $\epsilon_{i,(t-s,t-1)}$ as
\[
{\omega}^{\text{port}}_{i,t-1} = \mathrm{LongConv}_{\theta}(\epsilon_{i,(t-s,t-1)}),
\]
where $\theta$ denotes the learnable parameters of the LongConv model.
LongConv can capture complex time-series patterns. We chose it because of its linear scaling in the sequence length and simplicity. The choice of sequence model is flexible, and we expect alternative sequence models such as Transformers to perform similarly. Each convolution captures distinct time-series patterns, and our optimally tuned model has 32 different convolutions.

  \paragraph{Arbitrage Trading}
 The arbitrage portfolio return, $R^{\text{port}}_t$, is
 $$R^{\text{port}}_t = \epsilon_t^\top {\omega}^{\text{port}}_{t-1} = R_t^{\top} \left((\omega^{\epsilon}_{t-1} )^\top  {\omega}^{\text{port}}_{t-1} \right) = R_t^\top {\omega}_{t-1},$$
 with portfolio weights ${\omega}_{t-1} = (\omega^{\epsilon}_{t-1} )^\top  {\omega}^{\text{port}}_{t-1}$ in the asset space. Note that $\omega^{\epsilon}_{t-1}$ is only a function of lagged firm characteristics $X_{t-1}$, while $\omega^{\text{port}}_{t-1}$ is only a function of the time-series patterns in residuals. 

 \paragraph{Arbitrage Trading Objective}
We estimate the arbitrage portfolio weights to maximize the Sharpe ratio after transaction costs. 
We measure transaction costs as in \cite{dlsa}, which is common in this literature:
 $$\text{cost}(\omega_t, \omega_{t-1}) = 0.0005\times ||\omega_{t} - \omega_{t-1} ||_1 + 0.0001 \times ||max(-\omega_t,0)||_1,$$ 
The first penalty represents a transaction cost of 5 basis points per transaction, whereas the second one is a shorting cost of 1 basis point. 
The portfolio net return is then calculated as:
 $$R^{\text{port}}_{t,net} = R^{\text{port}}_t - \text{cost}(\omega_t, \omega_{t-1}).$$
Our objective function maximizes the net Sharpe ratio of the arbitrage portfolio and the explained variance of the factors. The tradeoff between these two objectives is selected optimally on the validation data. Including the explained variance is necessary for identification, and empirically improves the performance. This framework nests conditional latent factors that maximize explained variance as a special case.
\begin{align*}
\max_{\omega^F, \omega^{\text{port}}}  \underbrace{\frac{\bar{R}_{net}^{\text{port}} - R_f}{\sqrt{\frac{1}{T} \sum_{t=1}^{T} (R_{t,net}^{\text{port}} - \bar{R}_{net}^{\text{port}})^2}}}_{\text{net Sharpe ratio}} + \;\; \lambda_{\text{Var}} \cdot \underbrace{\frac{1}{N}\sum_{i=1}^N\left(1 - \frac{\operatorname{Var}(e^i)}{\operatorname{Var}(R^i)}\right)}_{\text{explained variance}},
\end{align*}
subject to $\| \omega_t \|_1=1$ and where $\bar{R}_{net}^{\text{port}} = \frac{1}{T} \sum_{t=1}^{T} R_{t,net}^{\text{port}}$, and $R_f$ is the risk-free rate. The learned parameters that determine $\omega^F$ and $\omega^{\text{port}}$ are the query matrix $Q$, the embedding matrix $W^K$, and the LongConv model parameters $\theta$.

\section{Empirical Analysis}

\subsection{Data}

\paragraph{Data Sets} We collect daily equity return data for the securities on CRSP from January 1990 through December 2021. Our analysis uses only the most liquid stocks. More specifically, we consider only the 500 largest stocks based on the previous month market capitalization. We complement the stock returns with 39 firm-specific characteristics from \citep{dlap}, which are listed in Table \ref{tab:firm_chars}. All these variables are constructed either from accounting variables from the CRSP/Compustat database or from past returns from CRSP. The full details on the construction of these variables are in the Internet Appendix of \citep{dlap}. Firm characteristics are normalized to rank quantiles as it is standard in this literature. In addition to the most important characteristics from \citep{dlap} we also include the previous day and week return and volatility. In order to keep the level information, we also include the cross-sectional median of the characteristics and the risk free rate, which results in the dimension of $X_{t-1}$ of 79. Missing values in the characteristics are imputed with last observed values if available and by the cross-sectional median otherwise.

\begin{table*}[t]
  \centering
  \tcapfig{Out-of-sample Annualized Performance}
   \label{tab:oos_annualized_performance}
  \begin{tabular}{l|l|ccccccc}
    \toprule
    Model & K & SR & $\mu$ & $\sigma$ & $SR_{\text{net}}$ & $\mu_{\text{net}}$ & $\sigma_{\text{net}}$ & Beta  \\
    \midrule
    \multirow{6}{*}{\makecell[l]{ \textbf{Attention Factors}}}
    &1  & \textbf{3.05} & 14.45 & 4.74 & \textbf{1.68} & 7.94 & 4.72 & 0.05  \\
    &3  & \textbf{3.05} & 14.91 & 4.89 & \textbf{1.69} & 8.25 & 4.87 & 0.06  \\
    &5 & \textbf{2.92} & 14.21 & 4.87 & \textbf{1.58} & 7.66 & 4.85 & 0.07  \\
    &8   & \textbf{3.35} & 15.70 & 4.68 & \textbf{1.94} & 9.05 & 4.66 & 0.07  \\
    &15 & \textbf{3.81} & 16.66 & 4.37 & \textbf{2.25} & 9.78 & 4.35 & 0.06  \\
    &30  & \textbf{3.97} & 16.66 & 4.20 & \textbf{2.28} & 9.52 & 4.18 & 0.05  \\
    &100 & \textbf{4.52} & 16.45 & 3.64 & \textbf{2.19} & 7.93 & 3.62 & 0.05  \\
    \midrule
    \multirow{7}{*}{\makecell[l]{Parametric Benchmark\\ PCA  +  OU Thresh}}
    & 1 & 0.40 & 1.85 & 4.57 & -2.54 & -11.62 & 4.57 & 0.02  \\
    & 3 & 1.26 & 4.18 & 3.33 & -2.72 & -9.04 & 3.33 & 0.01  \\
    & 5 & 0.99 & 2.91 & 2.93 & -3.44 & -10.11 & 2.93 & 0.00 \\
    & 8 & 0.78 & 2.04 & 2.61 & -4.15 & -10.83 & 2.61 & 0.00  \\
    & 10 & 0.80 & 1.99 & 2.50 & -4.32 & -10.80 & 2.50 & 0.00  \\
    & 15 & 0.51 & 1.12 & 2.20 & -5.24 & -11.53 & 2.20 & 0.00  \\
    & 30  & 0.18 & 0.40 & 2.24 & -6.45 & -14.74 & 2.29 & -0.00  \\
    & 100 & -0.35 & -0.66 & 1.87 & -7.05 & -13.23 & 1.88 & -0.00  \\
    \midrule  
    \multirow{7}{*}{\makecell[l]{PCA  Factors \\ (Two-Step Approach)}}
    & 1 & 2.26 & 13.10 & 5.79 & 1.19 & 6.98 & 5.78 & 0.10  \\
    & 3 & 2.76 & 14.61 & 5.30 & 1.57 & 8.29 & 5.28 & 0.07  \\
    & 5 & 2.41 & 14.10 & 5.86 & 1.30 & 7.62 & 5.84 & 0.10  \\
    & 8 & 2.64 & 14.88 & 5.63 & 1.50 & 8.42 & 5.61 & 0.09  \\
    & 10 & 2.66 & 14.94 & 5.61 & 1.52 & 8.48 & 5.59 & 0.09  \\
    & 15 & 2.56 & 14.74 & 5.75 & 1.41 & 8.08 & 5.73 & 0.09  \\
    & 30 & 2.79 & 15.15 & 5.42 & 1.57 & 8.47 & 5.40 & 0.09  \\    
    & 100 & 2.66 & 14.36 & 5.40 & 1.44 & 7.75 & 5.38 & 0.09  \\
    \midrule    
    \multirow{1}{*}{Market}
    &- & 0.42 & 8.61 & 20.37 & 0.42 & 8.61 & 20.37 & 1.00  \\
    \bottomrule
  \end{tabular}
  \bnotefig[0.8\linewidth]{The table shows the out-of-sample arbitrage trading performance using different models (Jan. 1998- Dec. 2021). We report the results for our Attention Factor model, and for a two-step approach where the residual obtained from PCA factors. The portfolio weight functions is estimated with LongConv from the residuals of each factor model. For each model, $K$ denotes the number of factors. Parametric benchmark OU+Thresh is the parametric Ornstein-Uhlenbeck model with thresholding trading policy based on \cite{avelee} and implemented as in \cite{dlsa}. We use a lookback window of $30$ days of residual returns, and PCA factors are estimated on rolling window of $252$ days. $K$ denotes the number of factors. The Sharpe Ratio (SR), mean return ($\mu$ in \%), and standard deviation ($\sigma$) are annualized. "Net" metrics account for transaction and shorting costs. Beta denotes the market beta. The equally weighted market portfolio is provided for reference.}
\end{table*}

\subsection{Estimation} 

\paragraph{Models} 
We estimate our models on a rolling window of 8 years, where we retrain the models every year and evaluate them out-of-sample from January 1998 to December 2021. We compare our Attention Factor model to two natural benchmark models, which both estimate residuals to maximize explained variation. The parametric benchmark from the seminal work of \citep{avelee} corresponds to classical mean-reversion trading. The second benchmark estimates factors with PCA, but uses the same flexible convolution model for trading the residuals. It allows us to understand the importance of the one-step optimization and general functional form. Both sets of benchmark models have been found to perform strongly empirically.

 In summary, the three classes of models are 
\begin{enumerate}
    \item \textbf{Attention Factors}: The factors and arbitrage trading policy are learned in one-step with a trading objective including transaction costs. We consider 1, 3, 5, 8, 10, 15, 30, and 100 latent factors. 
       \item \textbf{PCA Factors}: The latent factors are estimated with PCA using the past 252 trading days. The residual portfolios weights $\omega^{port}$ are estimated with our LongConv model and the Sharpe ratio objective with trading costs. Hence, we allow the same flexibility for $\omega^{\text{port}}_{t-1}$ as in our attention model. This is essentially a benchmark in the spirit of \citep{dlsa}. 
    \item \textbf{PCA+OU Thresh}: We use PCA factors to estimate residuals and use a parametric portfolio weight based on an Ornstein-Uhlenbeck model with thresholding rule proposed in \citep{avelee}. We use the same implementation as in \citep{dlsa}.
\end{enumerate}
We report the out-of-sample annualized Sharpe ratio, average return, volatility and net results after subtracting transaction costs.

 \paragraph{Implementation}
 \label{implementation}
Adam ~\citep{loshchilov2017decoupled} was used as our optimizer. We use the last two years of the first training data to select tuning parameters and find that our results are robust to the tuning parameter selection. We follow the architectural choices of prior work, and test our model for a wide range of parameter choices. Our optimal Attention Factor model has a hidden dimension of $d_x=32$. The optimal LongConv sequence model has 1 layer with hidden dimension 32. The optimal weight on the variance is $\lambda_{\text{Var}}=100$. All sequence models use a look-back window of the past 30 daily residual returns. 

\begin{figure}[t]
    \centering
    \tcapfig{Cumulative Returns of Arbitrage Portfolios}
    \includegraphics[width=1\linewidth]{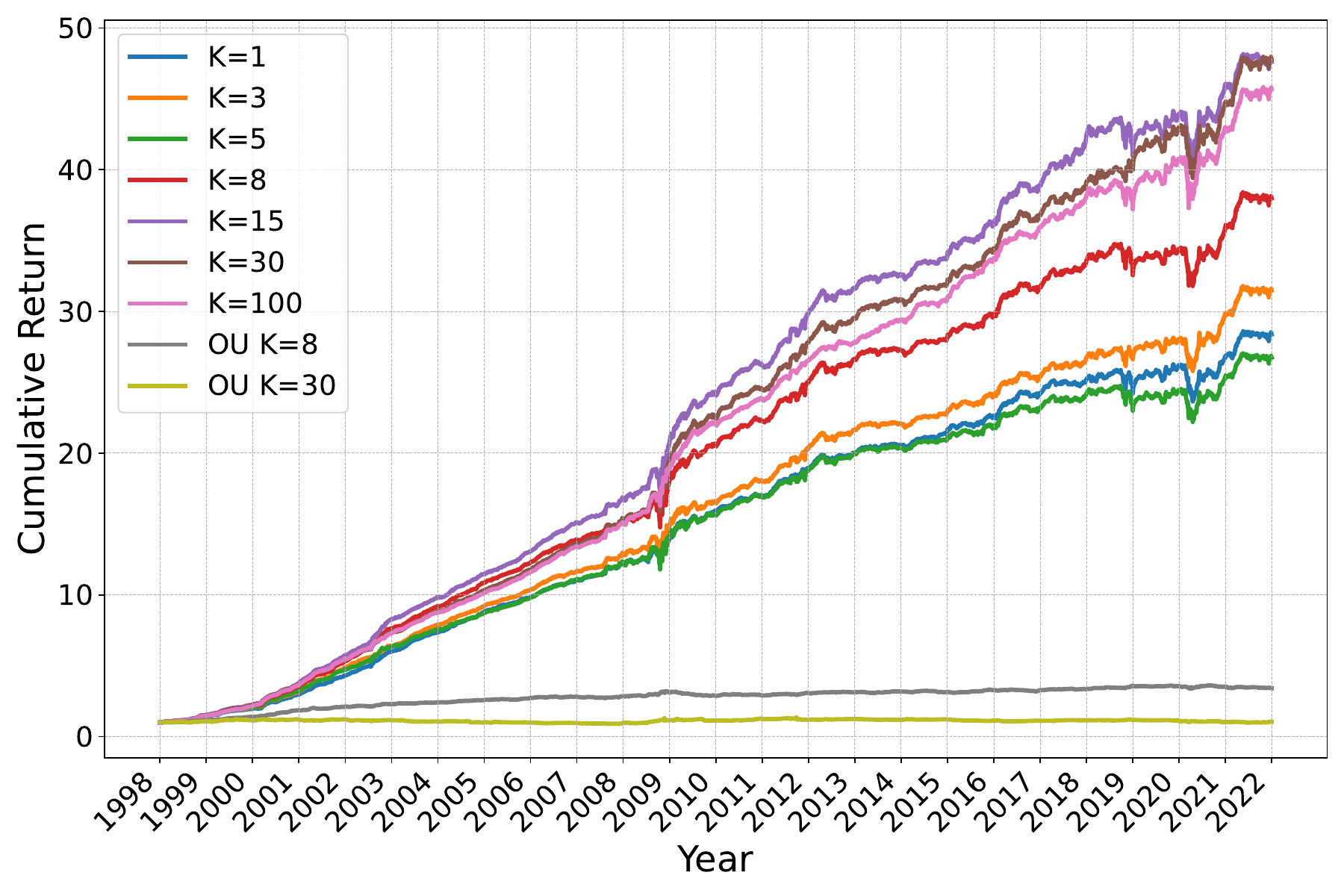}
    \bnotefig{The figure shows the out-of-sample cumulative returns for different arbitrage portfolios. We consider the Attention Factor model with different number of factors, K, and the parametric Ornstein-Uhlenbeck model with thresholding trading policy (OU) on the out-of-sample period from Jan 1998 to Dec 2021.}
    \label{fig:cum_returns}
\end{figure}

\subsection{Results}

 \paragraph{Performance}
 Table \ref{tab:oos_annualized_performance} shows the main results for our attention arbitrage model and the benchmarks for different number of factors. First, our Attention Factor model achieves an excellent performance as demonstrated by the annual Sharpe ratio of around 4 with 30 attention factors. It substantially outperforms the parametric benchmark, which can achieve a solid Sharpe ratio of 1.2. A two-step approach with PCA factors and the convolutional time-series filter achieves out-of-sample Sharpe ratios close to 2.8 with 30 PCA factors, which illustrates the importance of a flexible time-series filter. All models are essentially uncorrelated with a market factor. We conclude that our attention factors identify more profitable arbitrage opportunities. 

 \paragraph{Net Results} 
 Second, our attention factors represent the best performing model after transaction costs. We compare the Sharpe ratios and average returns after transaction costs. The strategy of the parametric model drops to negative net Sharpe ratios and mean returns due to its excessive turnover. Similarly, a simple PCA factor model cannot adjust the factor construction to trading frictions, resulting in a deterioration of the net performance to a Sharpe ratio of around 1.5. In contrast, our Attention Factor model provides unprecedented performance of 2.3 after trading costs. This is due to the end-to-end optimization and including the transaction costs in the objective function. The model learns to identify arbitrage strategy that is profitable after taking transaction costs into account. This makes our model one of the best performing models in the literature under realistic frictions.

 \paragraph{Number of Factors}
 Third, we demonstrate the importance of weak factors for arbitrage trading. Our attention arbitrage model achieves a substantial performance for 8 attention factors. However, we observe an out-of-sample improvement for including 30 factors. These higher order factors capture weak signals and local dependency patterns. This is in line with \citep{Lettau_Pelger_RFS}, who show that weaker factors that capture local dependency patterns are important for trading. A model with 100 attention factors leads to further minor improvements. This means that our model does not overfit, but discovers further weak signals, as increasing K expands the model's capacity to optimize the trading objective without requiring factor independence.

 \paragraph{Performance over Time} 
 Figure \ref{fig:cum_returns} shows cumulative out-of-sample returns for the different arbitrage strategies. Our attention factors exhibit a strong performance throughout the full sample - even during the later part of the sample where arbitrage trading is more challenging. The large-volatility period in early 2020 due to COVID-19 led to temporary deviations in market dynamics, reflecting a short-lived distribution shift relative to the preceding training windows. While a shorter training horizon may have been able to adapt more rapidly to these changing conditions, we maintain a fixed 8-year rolling window to ensure a consistent empirical design.

\begin{table*}[t]
  \centering
  \tcapfig{Characteristic Importance for Model Performance}
  \label{tab:feature_importance}
  \begin{tabular}{lcccccc}
    \toprule
    Dropped Feature & {SR} & $\mu$ & $\sigma$  & $SR_{\text{net}}$ & $\mu_{\text{net}}$ &  {Beta}  \\
    \midrule
   baseline   & \multicolumn{1}{c}{ \textbf{3.97} (0.13)} & 16.66 & 4.20 & \textbf{2.28} & 9.52  & 0.05    \\
   (none excluded) & & & & & & \\
    \midrule
past returns & \multicolumn{1}{c}{ 1.50 (0.07) } & 7.82 & 5.23 & 0.59 & 3.09 & 0.08  \\
investment & \multicolumn{1}{c}{ 3.88 (0.17) } & 17.93 & 4.63 & 2.19 & 10.06 & 0.06  \\
profitability & \multicolumn{1}{c}{ 3.94 (0.15) } & 18.39 & 4.67 & 2.26 & 10.48 & 0.05  \\
intangibles & \multicolumn{1}{c}{ 3.91 (0.15) } & 18.18 & 4.65 & 2.24 & 10.34 & 0.06  \\
value & \multicolumn{1}{c}{ 4.08 (0.12) } & 18.45 & 4.53 & 2.32 & 10.44 & 0.04  \\
trading frictions & \multicolumn{1}{c}{ 2.90 (0.14) } & 13.36 & 4.61 & 1.34 & 6.14 & 0.06  \\
    \bottomrule
  \end{tabular}
\bnotefig[0.8\linewidth]{The table shows the out-of-sample model performance when dropping characteristic groups in the estimation and evaluation. The Sharpe Ratio (SR), mean return ($\mu$ in \%), and standard deviation ($\sigma$) are annualized. "Net" metrics account for transaction and shorting costs. Reported Sharpe Ratio, mean return, and return standard deviation are annualized. Net Sharpe Ratio is calculated after accounting for transaction costs and shorting costs. Beta is relative to the market.
All values are averaged across multiple starting seeds for model estimation of the neural networks, and the value in parentheses for the Sharpe Ratio is the standard deviation across seeds. Each row drops a characteristic group to assess its feature group importance. The number of attention factors are 30. The out-of-sample evaluation is from Jan. 1998- Dec. 2021.
}
\end{table*}

\begin{figure*}[t]
    \centering
    \tcapfig{Interpretation of Attention Factor Betas}
    \includegraphics[width=0.85\linewidth]{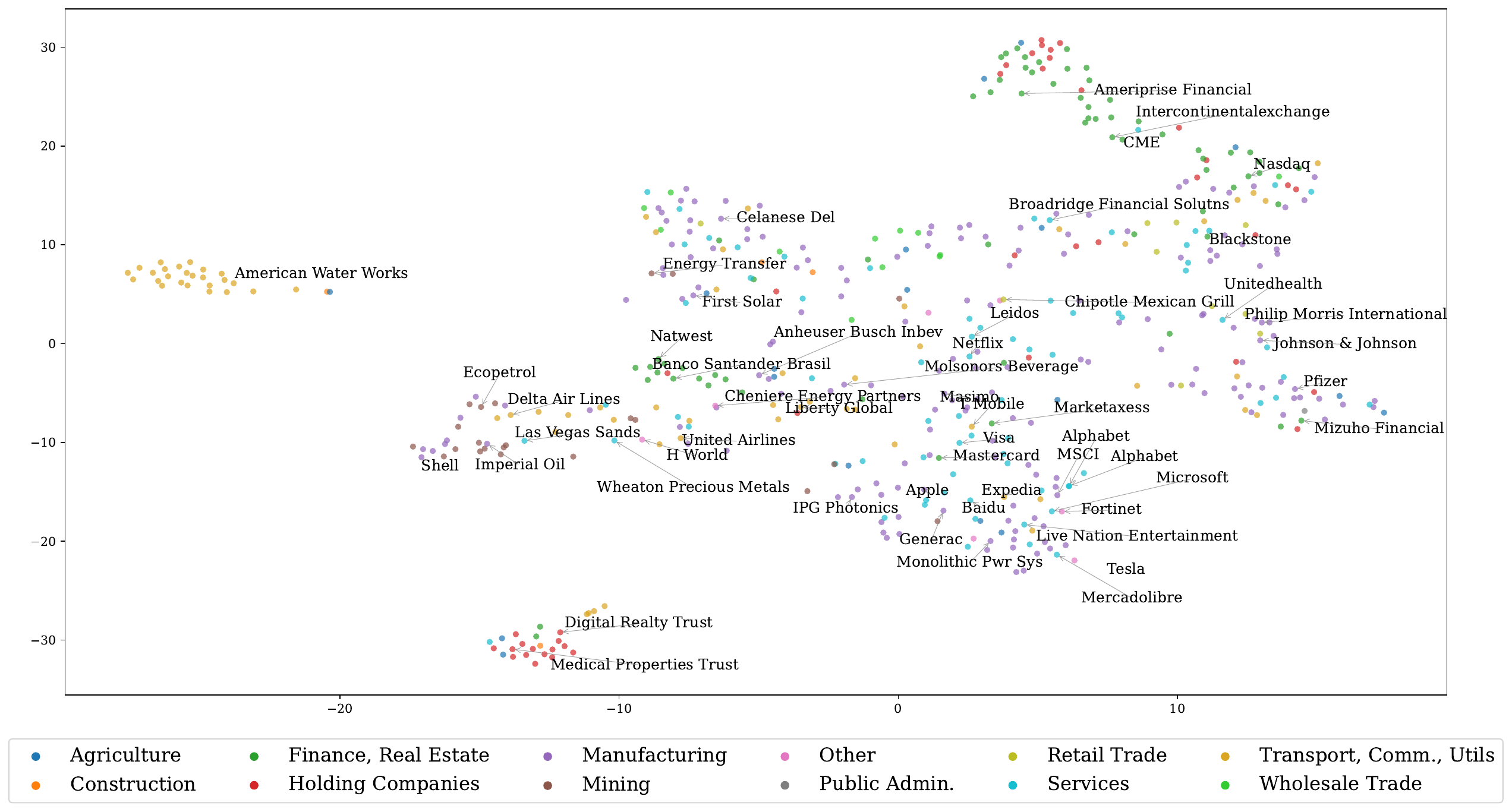}
    \bnotefig[0.8\linewidth]{The figure illustrates the attention factor loading composition by showing the t-SNE (t-distributed stochastic neighbor embedding) projection of estimated betas for the first 8 attention factors estimated on the 500 equities with the largest market cap on the first trading day of 2021. The training period is January 2013 - December 2020. The dots are colored based on industry classification. The Attention Factor model betas capture meaningful dependencies between firms. The clusters represent different industry sectors: the upper right cluster represents banks and financial firms,  lower right petroleum and energy companies, lower middle real estate companies, lower left utility companies, middle left hotels and middle-lower left technology.}
    \label{fig:tsne-beta}
\end{figure*}

\begin{figure*}[t]
    \centering
    \tcapfig{Attention Weights for Factor Portfolio Weights}
    \includegraphics[width=0.82\linewidth]{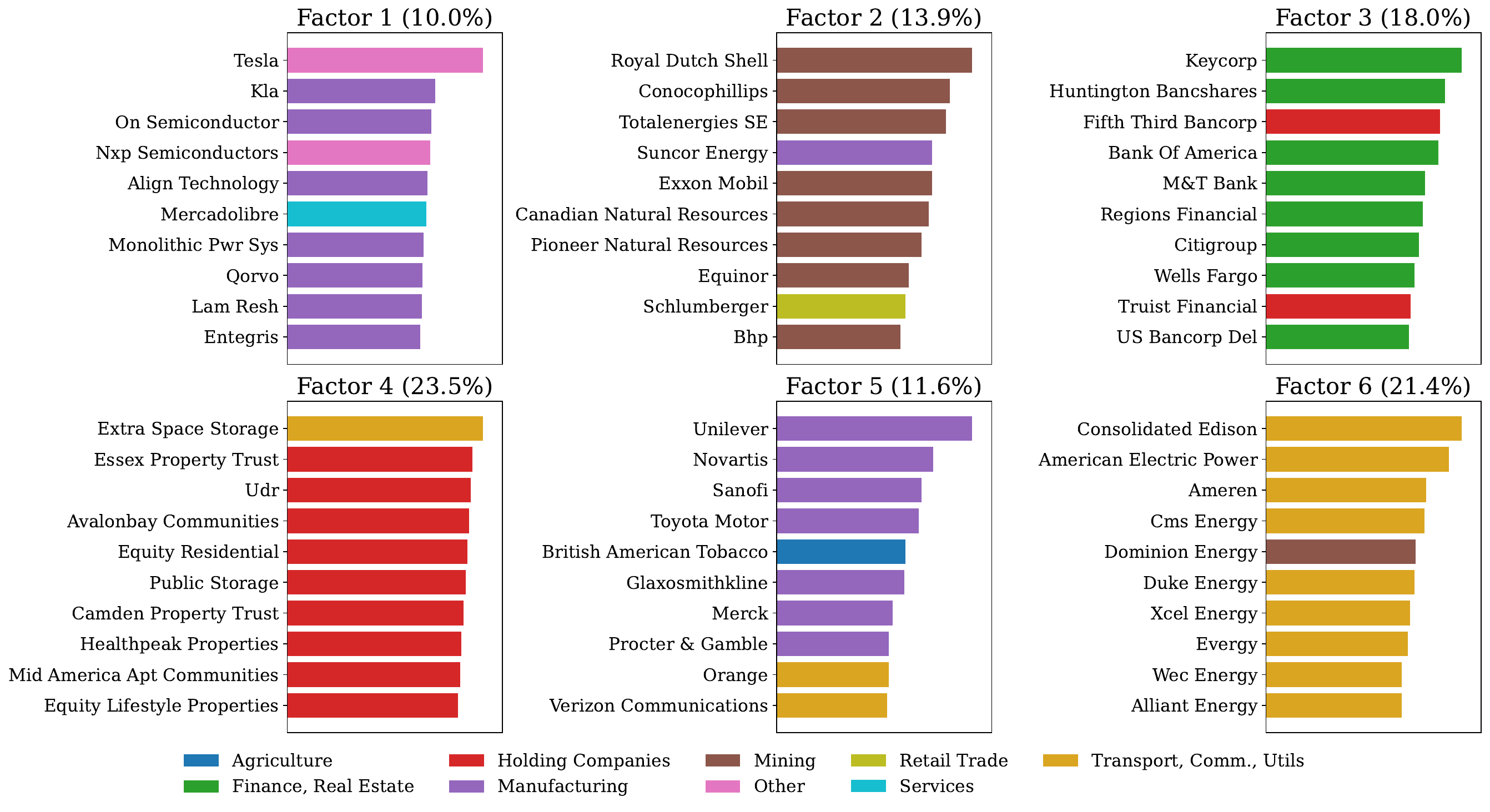}
    \bnotefig[0.8\linewidth]{This figure shows attention weights for constructing the factor portfolio weights. We display the top 10 company weights for the first 6 factors (out of total 8) for the Attention Factor model. We also report the industry for each company, along with the percentage weight (out of the full 500 assets) that the top 10 companies represent. The evaluation is on the first trading day of 2021 for the model trained from January 2013 to December 2020.}
    \label{fig:factor_constituents}
\end{figure*}

\subsection{Interpretation}

\paragraph{Drivers of Performance} 
Table \ref{tab:feature_importance} reports the effect on out-of-sample performance after removing a group of characteristics. Removing past return information in the attention factors substantially reduces the performance and the net Sharpe ratios drop to 0.59. In contrast, removing any other characteristic group has a negligible effect. This indicates that price based patterns and not ``classical'' firm characteristics are driving arbitrage trading. The values are averaged across multiple starting seeds for the neural network estimation, and the value in parentheses for the Sharpe Ratio is the standard deviation across seeds.
We conclude that the results are robust to the implementation.

\paragraph{Factor Structure} 
Our attention factor structure has a clear economic interpretation. We focus on the attention 8-factor model, which already achieves a substantial performance. Figure \ref{fig:tsne-beta} represents the similarity of firms captured by the loadings, that is, firms with similar loadings are considered closer. We use t-SNE to represent closeness of firms in the loading space. Here we evaluate our model in a specific date, but similar results hold throughout our sample. Firms in similar industries are grouped together, that is, our model learns specific industries. For example, the upper right cluster represents banks and financial, the lower right petroleum and energy companies, lower middle real estate companies, lower left utility companies, middle left hotels and middle-lower left technology companies. Note, we do not provide industry classification, but this similarity is learned from price data and firm fundamentals.

\paragraph{Factors Weights} 
The factor portfolio weights have a clear interpretation in terms of industry sectors. Figure \ref{fig:factor_constituents} shows the top 10 companies that are used to construct the first six factors on a representative date. In each case, the first top 10 companies account for a large portion of the total weight in the companies (between 10\%-23\%). We see clear industry relationships. Factor 1 represents technology, factor 2 natural resources, factor 3 the financial industry, factor 4 holding companies, factor 5 consumer manufacturing and factor 6 captures energy companies.

\section{Conclusion}

This paper develops an Attention Factor model for statistical arbitrage. We provide a one-step estimation framework for latent factors to identify similar assets and an arbitrage portfolio allocation based on time-series patterns. 
 The two key innovations are the conditional latent attention factors to capture complex dependencies in firm characteristics and the one-step model estimation that maximizes portfolio performance after trading frictions. In extensive empirical analysis, we demonstrate that our arbitrage model sets the new standard in this literature with the best performance under realistic trading frictions.

\bibliographystyle{ACM-Reference-Format}
\bibliography{references}
\balance
\appendix

\section{Sequence Model Details}
We use LongConv~\citep{fu2023simplehardwareefficientlongconvolutions} as the sequence model on the residual time series. 
Given input to the LongConv layer $u \in \mathbb{R}^{N\times d\times T}$ with $N$ assets, hidden dimension $d$, and sequence length $T$, 
a learnable long convolution kernel $\mathcal{K} \in \mathbb{R}^{d\times T}$ and skip parameter $D \in \mathbb{R}^{d}$, the LongConv computes
\[
y = \mathcal{K} * u + D \odot u,
\]
where $*$ denotes a convolution along the temporal dimension, and $\odot$ denotes element-wise multiplication.
The convolution is defined as
\[
(\mathcal{K} * u)[i] = \sum_j u[j]\,\mathcal{K}[i-j],
\]
which has a direct computational complexity of $\mathcal{O}(T^2)$ for a sequence of length $T$. 
In practice, we compute it efficiently using the FFT convolution theorem:
\[
\mathcal{K} * u = \mathcal{F}^{-1}\!\big(\mathcal{F}u \odot \mathcal{F}\mathcal{K}\big),
\]
which reduces the complexity to $\mathcal{O}(T\log T)$. 
Here, $\mathcal{F}$ and $\mathcal{F}^{-1}$ denote the discrete Fourier transform and its inverse.

\paragraph{Kernel regularization}
Following \citep{fu2023simplehardwareefficientlongconvolutions}, we apply the element-wise \emph{Squash} operator to $\mathcal{K}$ as a simple regularizer in the model forward pass:
\[
\bar{\mathcal{K}} = \mathrm{sign}(\mathcal{K})\odot \max(|\mathcal{K}|-\lambda_{\text{squash}},0),
\]
with regularization strength $\lambda_{\text{squash}}$. This acts as a proximal step for an $\ell_1$ penalty and regularizes the kernel by shrinking all weights and setting small weights to zero, making the kernel more sparse.

\paragraph{Initialization }
The kernel is initialized with a geometric decay across both the sequence and hidden dimensions following \citep{fu2023simplehardwareefficientlongconvolutions}:
\[
\mathcal{K}^{(h)}_t = x\,\exp\!\left(-\frac{t}{T}\,{(\frac{d}{2})}^{\frac{h}{d}}\right), 
\qquad x \sim \mathcal{N}(0,1),
\]
for $1 \leq t \leq T$ and $1 \leq h \leq d$, which gives convolution filters that act on both short and long time-scales.

 \section{Training Details}
 \paragraph{Model parameters}
 We use the last two years of the first training window to select tuning parameters, and find that our results are robust to the tuning parameter selection. The selected optimal parameters are shown in Table~\ref{tab:hyperparams}.

\begin{table}[h]
\tcapfig{Selected tuning parameters}
\label{tab:hyperparams}
\centering
\resizebox{\columnwidth}{!}{%
\begin{tabular}{l c p{5cm}}
\toprule
\textbf{Parameter} & \textbf{Value} & \textbf{Description} \\
\midrule
Hidden dim ($d$)           & 32 & Model hidden dimension \\
Dropout                & 0.1 & Model dropout\\ 
$d_x$         & 32 & Model
attention dim \\
Epochs                 & 30 & Number of passes over the data\\ 
Nr layers              & 1 & Number of layers for sequence model \\
$\lambda_{\text{VAR}}$            & 100 & Weight on variance term in loss  \\
LR & 0.003 & Learning rate\\
Weight decay & 0.05 & Adam weight decay in LongConv model \\
LongConv init & Geom Decay & Kernel initialization of the LongConv model\\
$\lambda_{\text{squash}}$ & 0.001 & LongConv Squash operator strength \\

\bottomrule
\end{tabular}
}
\end{table}   
 \paragraph{Computational Setup}
 All the results in the paper are obtained on a Linux cluster with 5 NVIDIA RTX A6000 GPUs, each with 49140 MB memory, running on CUDA Version 12.5. The cluster is equipped with two AMD EPYC 7763 64-Core Processors (128 physical cores, 256 threads total) and 1 TB of RAM. 

\end{document}